# Two Step CCA: A new spectral method for estimating vector models of words

**Paramveer S. Dhillon**  DHILLON@CIS.UPENN.EDU
Computer & Information Science, University of Pennsylvania, Philadelphia, PA 19104 U.S.A

**Jordan Rodu**  JRODU@WHARTON.UPENN.EDU
Statistics, The Wharton School, University of Pennsylvania, Philadelphia, PA 19104 U.S.A

**Dean P. Foster**  FOSTER@WHARTON.UPENN.EDU
Statistics, The Wharton School, University of Pennsylvania, Philadelphia, PA 19104 U.S.A

**Lyle H. Ungar**  UNGAR@CIS.UPENN.EDU
Computer & Information Science, University of Pennsylvania, Philadelphia, PA 19104 U.S.A

## Abstract

Unlabeled data is often used to learn representations which can be used to supplement baseline features in a supervised learner. For example, for text applications where the words lie in a very high dimensional space (the size of the vocabulary), one can learn a low rank "dictionary" by an eigendecomposition of the word co-occurrence matrix (e.g. using PCA or CCA). In this paper, we present a new spectral method based on CCA to learn an *eigenword* dictionary. Our improved procedure computes two set of CCAs, the first one between the left and right contexts of the given word and the second one between the projections resulting from this CCA and the word itself. We prove theoretically that this two-step procedure has lower sample complexity than the simple single step procedure and also illustrate the empirical efficacy of our approach and the richness of representations learned by our Two Step CCA (TSCCA) procedure on the tasks of POS tagging and sentiment classification.

## 1. Introduction and Related Work

Over the past decade there has been increased interest in using unlabeled data to supplement the labeled data in semi-supervised learning settings. Such methods help overcome the inherent data sparsity and provide improved generalization accuracies in high dimensional domains such as NLP. Approaches like (Ando & Zhang, 2005; Suzuki & Isozaki, 2008) have been empirically very successful and have achieved excellent accuracies on a variety of text data. However, it is often difficult to adapt these approaches to use in conjunction with existing supervised text learning systems, as these approaches enforce a particular choice of model.

An increasingly popular alternative is to learn representational embeddings for words from a large collection of unlabeled data (typically using a generative model), and to use these embeddings to augment the feature set of a supervised learner. Embedding methods learn "dictionaries" in low dimensional spaces or over a small vocabulary size, unlike the traditional approach of working in the original high dimensional vocabulary space with only one dimension "on" at a given time. Note that the dictionary provides a low ($\sim 30 - 50$) dimensional real-valued vector for each word *type*; I.e., all mentions of "bank" have the same vector associated with them.

The embedding methods broadly fall into two main categories:

1. *Clustering Based Embeddings:* Clustering methods, often hierarchical, are used to group distributionally similar words based on their contexts e.g. Brown Clustering (Brown et al., 1992; Pereira et al., 1993).

2. *Dense Embeddings:* These methods learn dense,

low dimensional, real-valued embeddings. Each dimension of these representations captures latent information about a combination of syntactic and semantic word properties. They can either be induced using neural networks like CW embeddings (Collobert & Weston, 2008) and *Hierarchical log-linear* (HLBL) embeddings (Mnih & Hinton, 2007) or by an eigen-decomposition of the word co-occurrence matrix, e.g. *Latent Semantic Analysis/Latent Semantic Indexing* (LSA/LSI) (Dumais et al., 1988) and *Low Rank Multi-View Learning* (LR-MVL) (Dhillon et al., 2011).

Our main focus in this paper is on eigen-decomposition based methods, as they have been shown to be fast and scalable for learning from large amounts of unlabeled data (Turney & Pantel, 2010; Dhillon et al., 2011), have a strong theoretical grounding, and are guaranteed to converge to globally optimal solutions (Hsu et al., 2009). Particularly, we are interested in Canonical Correlation Analysis (CCA) based methods as:

- *Firstly*, unlike PCA or LSA based methods they are scale invariant.

- *Secondly*, unlike LSA they can capture multi-view information. In text applications the left and right contexts of the words provide a natural split into two views which is totally ignored by LSA as it throws the entire context into a bag of words while constructing the term-document matrix.

Our main contributions in this paper are two fold. *Firstly*, we provide an improved method for learning an *eigenword dictionary* from unlabeled data using CCA – Two Step CCA (TSCCA). TSCCA computes two set of CCAs, the first one between the left and right contexts of the given word and the second one between the projections resulting from this CCA and the word itself. We prove theoretically that this two-step procedure has lower sample complexity than the simple single step procedure and also illustrate the empirical efficacy of our approach on the tasks of POS tagging and sentiment classification. *Secondly*, we show empirically that the dictionaries learned using CCA capture richer and more discriminative information than PCA or LSA on the same context, due to the fact that PCA and LSA are scale dependent and LSA further ignores the word order and hence the multi-view nature of context.

The remainder of the paper is organized as follows. In the next section we give our problem formulation and give a brief overview of CCA, which forms the core of our method. Section 3 describes our proposed two step CCA (TSCCA) algorithm in detail and gives theory supporting its performance. Section 4 demonstrates the effectiveness of TSCCA on tasks of POS tagging and sentiment classification. We conclude with a brief summary in Section 5.

## 2. Problem Formulation

Our goal is to estimate a vector for each word *type* that captures the distributional properties of that word in the form of a low dimensional representation of the correlation between that word and the words in its immediate context.

More formally, assume a document (in practice a concatenation of a large number of documents) consisting of $n$ tokens $\{\mathbf{w_1}, \mathbf{w_2}, ..., \mathbf{w_n}\}$, each drawn from a vocabulary of $v$ words. Define the left and right contexts of each token $\mathbf{w_i}$ as the $h$ words to the left or right of that token. The context sits in a very high dimensional space, since for a vocabulary of size $v$, each of the $2h$ words in the combined context requires an indicator function of dimension $v$. The tokens themselves sit in a $v$ dimensional space of words which we want to project down to a $k$ dimensional state space. We call the mapping from word types to their latent vectors the *eigenword dictionary*.

For a set of documents containing $n$ tokens, define $\mathbf{L_{n \times vh}}$ and $\mathbf{R_{n \times vh}}$ as the matrices specifying the left and right contexts of the tokens, and $\mathbf{W_{n \times v}}$ as the matrix of the tokens themselves. In $\mathbf{W}$, we represent the presence of the $j^{th}$ word type in the $i^{th}$ position in a document by setting matrix element $\mathbf{w_{ij}} = \mathbf{1}$. $\mathbf{L}$ and $\mathbf{R}$ are similar, but have columns for each word in each position in the context. (For example, in the sentence "I ate green apples yesterday.", for a context of size $h = 2$, the left context of "green" would be "I ate" and the right context would be "apples yesterday" and the third row of $\mathbf{W}$ would have a "1" in the column corresponding to the word "green".)

Define the complete context matrix $\mathbf{C}$ as the concatenation $[\mathbf{L}\ \mathbf{R}]$. Thus, for a trigram representation with vocabulary size $v$ words, history size $h = 1$, $\mathbf{C}$ has $2v$ columns – one for each possible word to the left of the target word and one for each possible word to the right of the target word.

$\mathbf{A_{cw}} = \mathbf{C}^\top \mathbf{W}$ then contains the counts of how often each word $\mathbf{w}$ occurs in each context $\mathbf{c}$, the matrix $\mathbf{A_{cc}} = \mathbf{C}^\top \mathbf{C}$ gives the covariance of the contexts, and $\mathbf{A_{ww}} = \mathbf{W}^\top \mathbf{W}$, the word covariance matrix, is a diagonal matrix with the counts of each word on the

diagonal.[1]

We want to find a vector representation of each of the $v$ word types such that words that are distributionally similar (ones that have similar contexts) have similar state vectors. We will do this using Canonical Correlation Analysis (CCA) (Hotelling, 1935; Hardoon & Shawe-Taylor, 2008), by taking the CCA between the combined left and right contexts $\mathbf{C} = [\mathbf{L}\ \mathbf{R}]$ and their associated tokens, $\mathbf{W}$.

CCA (Hotelling, 1935) is the analog to Principal Component Analysis (PCA) for pairs of matrices. PCA computes the directions of maximum covariance between elements in a single matrix, whereas CCA computes the directions of maximal correlation between a pair of matrices. Unlike PCA (and its variant LSA), CCA does not depend on how the observations are scaled. This invariance of CCA to linear data transformations allows proofs that keeping the dominant singular vectors (those with largest singular values) will faithfully capture any state information (Kakade & Foster, 2007). Secondly, CCA extends more naturally than LSA to sequences of words.[2] Remember that LSA uses "bags of words", which are good for capturing topic information, but fail for problems like part of speech (POS) tagging which need sequence information. Finally, as we show in the next section, the CCA formulation can be naturally extended to a two step procedure that, while equivalent in the limit of infinite data, gives higher accuracies for finite corpora.

More specifically, given a set of $n$ paired observation vectors $\{(l_1, r_1), ..., (l_n, r_n)\}$–in our case the two matrices are the left ($\mathbf{L}$) and right ($\mathbf{R}$) context matrices of a word–we would like to simultaneously find the directions $\mathbf{\Phi}_l$ and $\mathbf{\Phi}_r$ that maximize the correlation of the projections of $\mathbf{L}$ onto $\mathbf{\Phi}_l$ with the projections of $\mathbf{R}$ onto $\mathbf{\Phi}_r$. This is expressed as

$$\max_{\mathbf{\Phi}_l, \mathbf{\Phi}_r} \frac{\mathbb{E}[\langle \mathbf{L}, \mathbf{\Phi}_l \rangle \langle \mathbf{R}, \mathbf{\Phi}_r \rangle]}{\sqrt{\mathbb{E}[\langle \mathbf{L}, \mathbf{\Phi}_l \rangle^2] \mathbb{E}[\langle \mathbf{R}, \mathbf{\Phi}_r \rangle^2]}} \quad (1)$$

where $\mathbb{E}$ denotes the empirical expectation. We use the notation $\mathbf{C_{lr}}$ ($\mathbf{C_{ll}}$) to denote the cross (auto) covariance matrices between $\mathbf{L}$ and $\mathbf{R}$ (i.e. $\mathbf{L}^\top \mathbf{R}$ and $\mathbf{L}^\top \mathbf{L}$ respectively.).

The left and right canonical correlates are the solutions

---
[1] We will pretend that the means are all in fact zero and refer to these $\mathbf{A_{cc}}$ etc. as covariance matrices, when in fact they are second moment matrices.

[2] It is important to note that it is possible to come up with PCA variants which take sequence information into account, for instance by finding principal components of the $\mathbf{A_{cw}}$ matrix.

$\langle \mathbf{\Phi}_l, \mathbf{\Phi}_r \rangle$ of the following equations:

$$\mathbf{C_{ll}}^{-1} \mathbf{C_{lr}} \mathbf{C_{rr}}^{-1} \mathbf{C_{rl}} \mathbf{\Phi}_l = \lambda \mathbf{\Phi}_l$$
$$\mathbf{C_{rr}}^{-1} \mathbf{C_{rl}} \mathbf{C_{ll}}^{-1} \mathbf{C_{lr}} \mathbf{\Phi}_r = \lambda \mathbf{\Phi}_r \quad (2)$$

We keep the $k$ left and right singular vectors ($\mathbf{\Phi_l}$ and $\mathbf{\Phi_r}$) corresponding to the $\mathbf{k}$ largest singular values.

Using the above, we can define a "One step CCA" (OSCCA), procedure to estimate the *eigenword dictionary* as follows:

$$\mathbf{CCA}(\mathbf{C}, \mathbf{W}) \rightarrow (\mathbf{\Phi_C}, \mathbf{\Phi_W}) \quad (3)$$

where the $v \times k$ matrix $\mathbf{\Phi_W}$ contains the "eigenword dictionary" that characterizes each of the $v$ words in the vocabulary using a $k$ dimensional vector. More generally, the "state" vectors $\mathbf{S}$ for the $n$ tokens can be estimated either from the context as $\mathbf{C\Phi_C}$ or (trivially) from the tokens themselves as $\mathbf{W\Phi_W}$. Its important to note that both these estimation procedures give a redundant estimate of the same hidden "state."

The right canonical correlates found by OSCCA give an optimal approximation to the state of each word, where "optimal" means that it gives the linear model of a given size, $k$ that is best able to estimate labels that depend linearly on state, subject to only using the word and not its context. The left canonical correlates similarly give optimal state estimates given the context. See (Kakade & Foster, 2007) for more technical details, including the fact that these results are asymptotic in the limit of infinite data.

OSCCA, as defined in Equations 2 and 3 thus gives an efficient way to calculate the attribute dictionary $\mathbf{\Phi_W}$ for a set of $v$ words given the context and associated word matrices from a corpus. As mentioned, OSCCA is optimal only in the limit of infinite data. In practice, data is, of course, always limited. In languages, lack of data comes about in two ways. Some languages are resource poor; one just does not have that many tokens of them (especially languages that lack a significant written literature). Even for most modern languages, many of the individual words in them are quite rare. Due to the Zipfian distribution of words, many words do not show up very often. A typical year's worth of Wall Street Journal text only has "lasagna" or "backpack" a handful of times and "ziti" at most once or twice. To overcome these issues we propose a two-step CCA (TSCCA) procedure which has better sample complexity for rare words.

## 3. Two Step CCA (TSCCA) Algorithm

We now introduce our two step procedure (TSCCA) of computing an *eigenword dictionary* and show theoret-

**Algorithm 1** Two step CCA
1: **Input: L, W, R**
2: $\mathbf{CCA}(\mathbf{L},\mathbf{R}) \rightarrow (\mathbf{\Phi_L}, \mathbf{\Phi_R})$
3: $\mathbf{S} = [\mathbf{L\Phi_L} \;\; \mathbf{R\Phi_R}]$
4: $\mathbf{CCA}(\mathbf{S},\mathbf{W}) \rightarrow (\mathbf{\Phi_S}, \mathbf{\Phi_W})$
5: **Output: $\mathbf{\Phi_W}$**, the eigenword dictionary ically that it gives better estimates than the OSCCA method described above.

In the two-step method, instead of taking the CCA between the combined context [**L R**] and the words **W**, we first take the CCA between the left and right contexts and use the result of that CCA to estimate the state **S** of all the tokens in the corpus from their contexts. Note that we get partially redundant state estimates from the left context and from the right context; these are concatenated to make combined state estimate. This will contain some redundant information, but will not lose any of the differences in information from the left and right sides. We then take the CCA between **S** and the words **W** to get our final *eigenword dictionary*. This is summarized in Algorithm 1. The first step, the CCA between **L** and **R**, must produce at least as many canonical components as the second step, which produces the final output.

The two step method requires fewer tokens of data to get the same accuracy in estimating the *eigenword dictionary* because its final step estimates fewer parameters $O(vk)$ than the OSCCA does $O(v^2)$.

Before stating the theorem, we first explain this intuitively. Predicting each word as a function of all other word combinations that can occur in the context is far sparser than predicting low dimensional state from context, and then predicting word from state. Thus, for relatively infrequent words, OSCCA should have significantly lower accuracy than the two step version. Phrased differently, mapping from context to state and then from state to word (TSCCA) gives a more parsimonious model than mapping directly from context to word (OSCCA).

The relative ability of OSCCA to estimate hidden state compared to that of TSCCA can be summarized as follows:

**Theorem 1** *Given a matrix of words, **W** and their associated left and right contexts, **L** and **R** with vocabulary size $v$, context size $h$, and corpus of $n$ tokens. The ratio of the dimension of the hidden state that needs to be estimated by TSCCA in order to recover with high probability the information in the true state to the corresponding dimension needed for OS-CCA is $\frac{h+k}{hv}$.*

Please see the supplementary material for a proof of the above theorem.

Since the corpora we care about (i.e. text and language corpora) usually have $vh \gg h + k$, the TSCCA procedure will in expectation correctly estimate hidden state with a much smaller number of components $k$ than the one step procedure. Or, equivalently, for an estimated hidden state of given size $k$, TSCCA will correctly estimate more of the hidden state components.

As mentioned earlier, words have a Zipfian distribution so most words are rare. For such rare words, if one does a CCA between them and their contexts, one will have very few observations, and hence will get a low quality estimate of their eigenword vector. If, on the other hand, one first estimates a state vector for the rare words, and then does a CCA between this state vector and the context, the rare words can be thought of as borrowing strength from more common distributionally similar words. For example, "umbrage" (56,020) vs. "annoyance" (777,061) or "unmeritorious" (9,947) vs. "undeserving" (85,325). The numbers in parentheses are the number of occurrences of these words in the Google n-gram collection used in our experiments.

### 3.1. Practical Considerations

As mentioned in Section 2, CCA (either one-step or two-step) is essentially done by taking the singular value decomposition of a matrix. For small matrices, this can be done using standard functions in e.g. MATLAB, but for very large matrices (e.g. for vocabularies of tens or hundreds of thousands of words), it is important to take advantage of recent advances in SVD algorithms. For the experiments presented in this paper we use the method of (Halko et al., 2011), which uses random projections to compute SVD of large matrices.

## 4. Experimental Results

This section describes the performance (accuracy and richness of representation) of our *eigenword dictionary* learned via CCA. We evaluate the quality of the *eigenword dictionary* by using it in a supervised learning setting to predict a wide variety of labels that can be attached to words. For simplicity, all results shown here map from word *type* to label[3]; i.e. each word

---
[3] Its conceivable to learn eigenword dictionaries which map each token to a label, e.g. as done by (Dhillon et al., 2011) but that is not the focus of this paper.

| Language | Number of POS tags | Number of tokens |
|---|---|---|
| English | 17 (45) | 100311 |
| Danish | 25 | 100238 |
| Bulgarian | 12 | 100489 |
| Portuguese | 22 | 100367 |

*Table 1.* Description of the POS tagging datasets

type is assumed to have a single POS tag or type of sentiment.

We first compare the One-Step vs. Two Step CCA (TSCCA) procedures on a set of Part of Speech (POS) tagging problems for different languages, looking at how the predictive accuracy scales with corpus size for predictions on a fixed vocabulary. These results use small corpora. We then turn to the question of what types of semantic category information is captured by our *eigenword dictionaries* and how favorably its predictiveness compares with other state-of-the-art embeddings e.g. CW (Collobert & Weston, 2008) and HLBL (Mnih & Hinton, 2007). Here, we use the RCV1 Reuters newswire data as the unlabeled data.

### 4.1. POS Tagging: One step CCA (OSCCA) vs. Two step CCA (TSCCA)

We compare performance of OSCCA and TSCCA on the task of POS tagging in four different languages.

Table 1 provides statistics of all the corpora used, namely: the Wall Street Journal portion of the Penn treebank (Marcus et al., 1993) (we consider both the 17 tags of (PTB 17) (Smith & Eisner, 2005) and the 45 tags version of it (PTB 45)), the Bosque subset of the Portuguese Floresta Sinta(c)tica Treebank (Afonso et al., 2002), the Bulgarian BulTreeBank (Simov et al., 2002) (with only the 12 coarse tags), and the Danish Dependency Treebank (DDT) (Kromann, 2003).

Note that some corpora like English have $\sim 1$ million tokens whereas Danish only has $\sim 100k$ tokens. So, to address this data imbalance we kept only the first $\sim 100k$ tokens of the larger corpora so as to perform a uniform evaluation across all corpora.

Theorem 1 implies that the difference between OSCCA and TSCCA should be more pronounced at smaller sample sizes, where the errors are higher and that they should have similar predictive power asymptotically when we learn them using large amounts of data. So, we evaluate the performance of the methods on varying data sizes ranging from $5k$ to the entire $100k$ tokens. We take history size of $h = 1$ for CCA i.e. a word to the left and a word to the right; for PCA this reduces to a bag of trigrams. The PCA baseline used is similar to the one that has recently been proposed by (Lamar et al., 2010) except that here we are interested in supervised accuracy and not the unsupervised accuracy as in that paper. It is important to note that for POS tagging usually a trigram context (in our case $h = 1$) is sufficient to get state-of-the-art performance as can be substantiated by trigram POS taggers e.g. (Merialdo, 1994), so we need not consider longer contexts.

As mentioned earlier, for the unlabeled learning part i.e. learning using CCA/PCA we are interested in seeing the eigenword dictionary estimates for the word types (for a fixed vocabulary) get better with more data. So, when varying the unlabeled data from $5k$ to $100k$ we made sure that they had the exact same vocabulary and that the performance improvement is not coming from word types not present in the $5k$ tokens but present in the total $100k$.

To evaluate the predictive accuracy of the descriptors learned using different amounts of unlabeled data, we learn a multi-class SVM (Chang & Lin, 2001) with a linear kernel to predict the POS tag of each type. The SVM was trained using 80% of the word types chosen randomly and then tested on the remaining 20% types and this procedure was repeated 10 times. The hyperparameters of the linear SVM i.e. the cost function $C$ was chosen by cross validation on the training set. Its important to note that our train and test sets do not contain any of the same word types.[4] The value of $k$, the size of low dimensional projection was fixed at 50; The results were robust to the size of $k$ within the range of 30 to 100.

The accuracy of using OSCCA, TSCCA and PCA features in a supervised learner are shown in Figure 1 for the task of POS tagging. As can be seen from the results, CCA-based methods are significantly better (5% significance level in a paired t-test) than the PCA-based supervised learner. Among the CCAs, TSCCA is significantly better than OSCCA for small amounts of data, and (as predicted by theory) the two become comparable in accuracy as the amount of unlabeled data used to learn the CCAs becomes large.

### 4.2. Sentiment Classification Task: Richness of state learned by CCA

In the above results, we compared languages using part of speech tags, but the states estimated by CCA are far richer. We illustrate this by using them to build predictive models in English for a number of differ-ent word categories and also comparing them against other state-of-the-art embeddings e.g. CW (Collobert & Weston, 2008) and HLBL (Mnih & Hinton, 2007).

---
[4]We are doing non-disambiguating POS tagging i.e. each word type has a single POS tag, so if the same word type occurred in both the training and testing data, a learning algorithm that just memorized the training set would perform reasonably well.

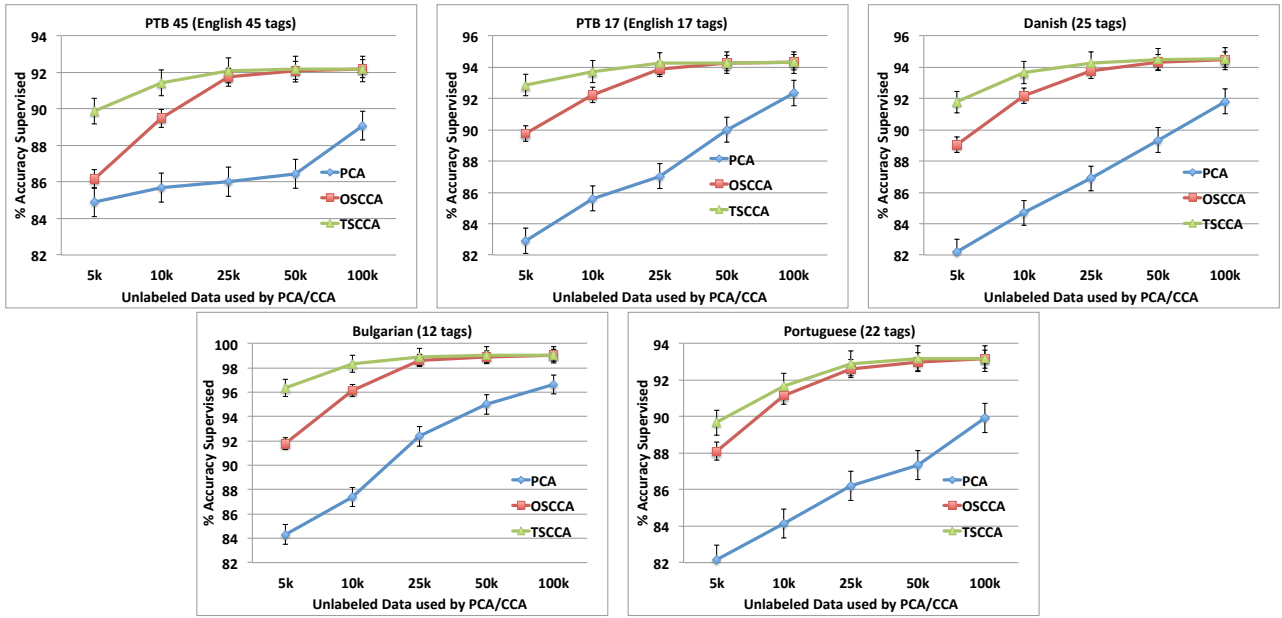

*Figure 1.* Plots showing accuracy as a function of number of tokens used to train the PCA/CCA for various languages.
**Note:** The results are averaged over 10 random, 80 : 20 splits of word types.

It is often useful to group words into semantic classes such as colors or numbers, professions or disciplines, happy or sad words, words of encouragement or discouragement, etc.

Many people have collected sets of words that indicate positive or negative sentiment. More generally, substantial effort has gone into creating hand-curated words that can be used to capture a variety of opinions about different products, papers, or people. For example (Teufel, 2010) contains dozens of carefully constructed lists of words that she uses to categorize what authors say about other scientific papers. Her categories include "problem nouns" (caveat, challenge, complication, contradiction,...), "comparison nouns" (accuracy, baseline, comparison, evaluation,...), "work nouns" (account, analysis, approach,ldots) as well as more standard sets of positive, negative, and comparative adjectives.

In the example below, we use words from a set of five dimensions that have been identified in positive psychology under the acronym PERMA (Seligman, 2011):
- *Positive emotion* (aglow, awesome, bliss, ...),
- *Engagement* (absorbed, attentive, busy, ...),
- *Relationships* (admiring, agreeable, ...),
- *Meaning* (aspire, belong, ...)
- *Achievement* (accomplish, achieve, attain, ...).

| Word sets | Number of observations | |
|---|---|---|
| | Class I | Class II |
| Positive emotion or not | 81 | 162 |
| Meaningful life or not | 246 | 46 |
| Achievement or not | 159 | 70 |
| Engagement or not | 208 | 93 |
| Relationship or not | 236 | 204 |

*Table 2.* Description of the datasets used. All the data was collected from the PERMA lexicon.

For each of these five categories, we have both positive words – ones that connote, for example, *achievement*, and negative words, for example, *un-achievement* (amateurish, blundering, bungling, ...). We would hope (and we show below that this is in fact true), that we can use our eigenword dictionary not only to distinguish between different PERMA categories, but also to address the harder task of distinguishing between positive and negative terms in the same category. (The latter task is harder because words that are opposites, such as "large" and "small," often are distributionally similar.)

The description of the PERMA datasets is given in Table 2. All of the following predictions use a single eigenword dictionary, which we estimated using TSCCA on the RCV1 corpus which contains Reuters newswire from Aug 96 to Aug 97 (about 63 million tokens in 3.3 million sentences). We scaled our TSCCA features to have a unit $\ell_2$ norm for each word type. As far as the other embeddings are concerned i.e. CW and HLBL, we downloaded them from http://metaoptimize.com/projects/wordreprs with k=50 dimensions and scaled as described in the paper (Turian et al., 2010).[5]

Figure 2 shows results for the five PERMA categories. The plots show accuracy as a function of the size of the training set used in the supervised portion of the learning. As earlier, we used SVM with linear kernel for the supervised binary classification, with the cost parameter chosen by cross-validation on training set and the value of $k$ was again fixed at 50.

As can be seen from the plots, the CCA variant TSCCA performs significantly (5% significance level in a paired t-test) better than PCA, CW and HLBL in 3 out of 5 cases and is comparable on the remaining 2 cases.

## 5. Conclusion

In this paper we proposed a new and improved spectral method, a two-step alternative (TSCCA) to the standard CCA (OSCCA) which can be used in domains such as Text/NLP which contain word sequences and where one has three views (the left context, the right context, and the words of interest themselves). We showed theoretically that the eigenword dictionaries learned by TSCCA provide more accurate state estimates (lower sample complexity) for small corpora than standard OSCCA. This was evidenced by superior empirical performance of TSCCA as compared to OSCCA and to simple PCA on the task of POS tagging, especially when less unlabeled data was used to learn the PCA or CCA representations.

We also showed empirically that the vector representations learned by CCA are much richer and contain more discriminative information than the representations learned by PCA as well as other state-of-the-art embeddings. Since CCA is scale-invariant and is able to take the multi-view nature of word sequence (i.e. the words to the left and words to the right) into account, it is able to learn more fine-grained spectral representations than PCA or LSA which ignore the word ordering.

## References

Afonso, S., Bick, E., Haber, R., and Santos, D. Floresta sinta(c)tica: a treebank for portuguese. In *In Proc. LREC*, pp. 1698–1703, 2002.

[5]Note that we did not validate the size of embeddings i.e. k, rather we just fixed it at 50 for all the methods.

Ando, R. and Zhang, T. A framework for learning predictive structures from multiple tasks and unlabeled data. *Journal of Machine Learning Research*, 6:1817–1853, 2005.

Brown, P., deSouza, P.., Mercer, R., Pietra, V. Della, and Lai, J. Class-based n-gram models of natural language. *Comput. Linguist.*, 18:467–479, December 1992. ISSN 0891-2017.

Chang, Chih-Chung and Lin, Chih-Jen. "LIBSVM: a library for support vector machines", 2001. Software available at http://www.csie.ntu.edu.tw/~cjlin/libsvm.

Collobert, R. and Weston, J. A unified architecture for natural language processing: deep neural networks with multitask learning. ICML '08, pp. 160–167, New York, NY, USA, 2008. ACM.

Dhillon, Paramveer S., Foster, Dean, and Ungar, Lyle. Multi-view learning of word embeddings via cca. In *Advances in Neural Information Processing Systems (NIPS)*, volume 24, 2011.

Dumais, S., Furnas, G., Landauer, T., Deerwester, S., and Harshman, R. Using latent semantic analysis to improve access to textual information. In *SIGCHI Conference on human factors in computing systems*, pp. 281–285. ACM, 1988.

Halko, Nathan, Martinsson, Per-Gunnar, and Tropp, Joel A. Finding structure with randomness: Probabilistic algorithms for constructing approximate matrix decompositions. *SIAM Rev*, 2011.

Hardoon, David and Shawe-Taylor, John. Sparse cca for bilingual word generation. In *EURO Mini Conference, Continuous Optimization and Knowledge-Based Technologies*, 2008.

Hotelling, H. Canonical correlation analysis (cca). *Journal of Educational Psychology*, 1935.

Hsu, D., Kakade, S., and Zhang, T. A spectral algorithm for learning hidden markov models. In *COLT*, 2009.

Kakade, S M. and Foster, Dean P. Multi-view regression via canonical correlation analysis. In Bshouty, Nader H. and Gentile, Claudio (eds.), *COLT*, volume 4539 of *Lecture Notes in Computer Science*, pp. 82–96. Springer, 2007.

Kromann, Matthias T. The danish dependency treebank and the underlying linguistic theory. in second workshop on treebanks and linguistic theories (tlt). In *In Proc. LREC*, pp. 217–220, 2003.

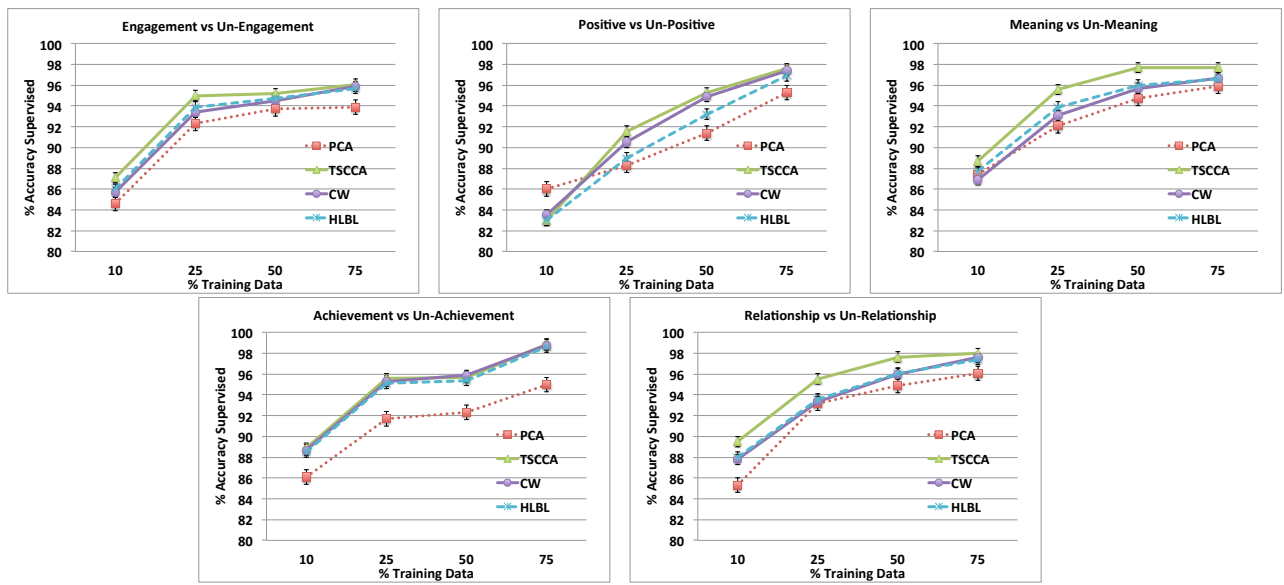

*Figure 2.* Test set prediction accuracies as a function of the amount of labeled data used in the supervised training (SVM) comparing TSCCA against PCA, CW and HLBL for distinguishing between positive and negative valences of each of the five PERMA categories. **Note:** 1). The results are averaged over 10 random splits. 2). The accuracies obtained by all methods were far higher than majority label baseline, so we omitted it from the plot.

Lamar, Michael, Maron, Yariv, Johnson, Mark, and Bienenstock, Elie. Svd and clustering for unsupervised pos tagging. ACL Short '10, pp. 215–219, Stroudsburg, PA, USA, 2010. Association for Computational Linguistics.

Marcus, Mitchell P., Marcinkiewicz, Mary Ann, and Santorini, Beatrice. Building a large annotated corpus of english: the penn treebank. *Comput. Linguist.*, 19:313–330, June 1993. ISSN 0891-2017.

Merialdo, Bernard. Tagging english text with a probabilistic model. *Comput. Linguist.*, 20, jun 1994.

Mnih, A. and Hinton, G. Three new graphical models for statistical language modelling. ICML '07, pp. 641–648, New York, NY, USA, 2007. ACM. ISBN 978-1-59593-793-3.

Pereira, F., Tishby, N., and Lee, L. Distributional clustering of English words. In *31st Annual Meeting of the ACL*, pp. 183–190, 1993.

Seligman, Martin. *Flourish: A Visionary New Understanding of Happiness and Well-being.* Free Press, 2011.

Simov, K., Osenova, P., Slavcheva, M., Kolkovska, S., Balabanova, E., Doikoff, D., Ivanova, K., Simov, A., Simov, E., and Kouylekov, M. Building a linguistically interpreted corpus of bulgarian: the bultreebank. In *In Proc. LREC*, 2002.

Smith, Noah A. and Eisner, Jason. Contrastive estimation: training log-linear models on unlabeled data. ACL '05, pp. 354–362. Association for Computational Linguistics, 2005.

Suzuki, J. and Isozaki, H. Semi-supervised sequential labeling and segmentation using giga-word scale unlabeled data. In *In ACL*, 2008.

Teufel, Simone. *The structure of scientific articles.* CSLI Publications, 2010.

Turian, J., Ratinov, L., and Bengio, Y. Word representations: a simple and general method for semi-supervised learning. ACL '10, pp. 384–394, Stroudsburg, PA, USA, 2010. ACL.

Turney, P.D. and Pantel, P. From frequency to meaning: vector space models of semantics. *Journal of Artificial Intelligence Research*, 37:141–188, 2010.